\documentclass[10pt,twocolumn,letterpaper]{article}

\usepackage[pagenumbers]{cvpr} 

\usepackage{graphicx}
\usepackage{amsmath}
\usepackage{amssymb}
\usepackage{booktabs}

\usepackage{comment}

\usepackage{multicol}
\usepackage{multirow}
\usepackage{bm}

\usepackage[accsupp]{axessibility}

\usepackage[pagebackref,breaklinks,colorlinks]{hyperref}

\usepackage[capitalize]{cleveref}
\crefname{section}{Sec.}{Secs.}
\Crefname{section}{Section}{Sections}
\Crefname{table}{Table}{Tables}
\crefname{table}{Tab.}{Tabs.}

\def\cvprPaperID{3623} 
\def\confName{CVPR}
\def\confYear{2022}

\newcommand{\x}{{\bm{x}}}

\newcommand{\z}{{\bm{z}}}
\renewcommand{\o}{{\bm{o}}}

\newcommand{\btheta}{{\bm{\theta}}}

\begin{document}

\title{An Image Patch is a Wave: Phase-Aware Vision MLP}

\author{
	Yehui Tang\textsuperscript{\rm 1,2},
	Kai Han\textsuperscript{\rm 2}, 
	Jianyuan Guo\textsuperscript{\rm 2,3}, 
	Chang Xu\textsuperscript{\rm 3},\\
	Yanxi Li\textsuperscript{\rm 2,3}, 
	Chao Xu\textsuperscript{\rm 1},	
	Yunhe Wang\textsuperscript{\rm 2}\thanks{Corresponding author.} \\
	\textsuperscript{\rm 1}School of Artificial Intelligence, Peking University. 
	\textsuperscript{\rm 2}Huawei Noah’s Ark Lab. \\
	\textsuperscript{\rm 3}School of Computer Science,  University of Sydney.\\
	yhtang@pku.edu.cn, \{kai.han, yunhe.wang\}@huawei.com.
}
\maketitle

\begin{abstract}

In the field of computer vision, recent works show that a pure MLP architecture mainly
stacked by fully-connected layers can achieve competing performance with CNN and transformer. An input image of vision MLP is usually split into multiple tokens (patches), while the existing MLP models directly aggregate them with fixed weights, neglecting the varying semantic information of tokens from different images. To dynamically aggregate tokens, we propose to represent each token as a wave function with two parts, amplitude and phase. Amplitude is the original feature and the phase term is a complex value changing according to the semantic contents of input images. Introducing the phase term can dynamically modulate the relationship between tokens and fixed weights in MLP. Based on the wave-like token representation, we establish a novel Wave-MLP architecture for vision tasks. Extensive experiments demonstrate that the proposed Wave-MLP is superior to the state-of-the-art MLP architectures on various vision tasks such as image classification, object detection and semantic segmentation. The source code is available at \url{https://github.com/huawei-noah/CV-Backbones/tree/master/wavemlp_pytorch} and \url{https://gitee.com/mindspore/models/tree/master/research/cv/wave_mlp}.

\end{abstract}

\section{Introduction}
\label{sec:intro}

In computer vision, convolutional neural networks (CNNs) have been the mainstream architectures for a long time~\cite{krizhevsky2012imagenet,he2016deep,radosavovic2020designing}. It is challenged by the recent works~\cite{dosovitskiy2020image,wang2021pyramid,liu2021Swin}, in which a standard Transformer~\cite{vaswani2017attention} model can also work well on various computer vision tasks, such as image classification, object detection and semantic segmentation~\cite{han2020survey}. Considering the high complexity of self-attention modules in the vision transformer,  more simple architectures (\eg, MLP-Mixer~\cite{tolstikhin2021mlp}, ResMLP\cite{touvron2021resmlp}) stacking only multi-layer perceptrons (MLPs) have attracted much attention.  Compared with CNNs and Transformers, these vision  MLP architectures involve less inductive bias and have potential to be applied on more diverse tasks.

\begin{figure}[h]
	\centering
	\small
	\begin{subfigure}{0.48\linewidth}
		
		\includegraphics[width=1.0\linewidth]{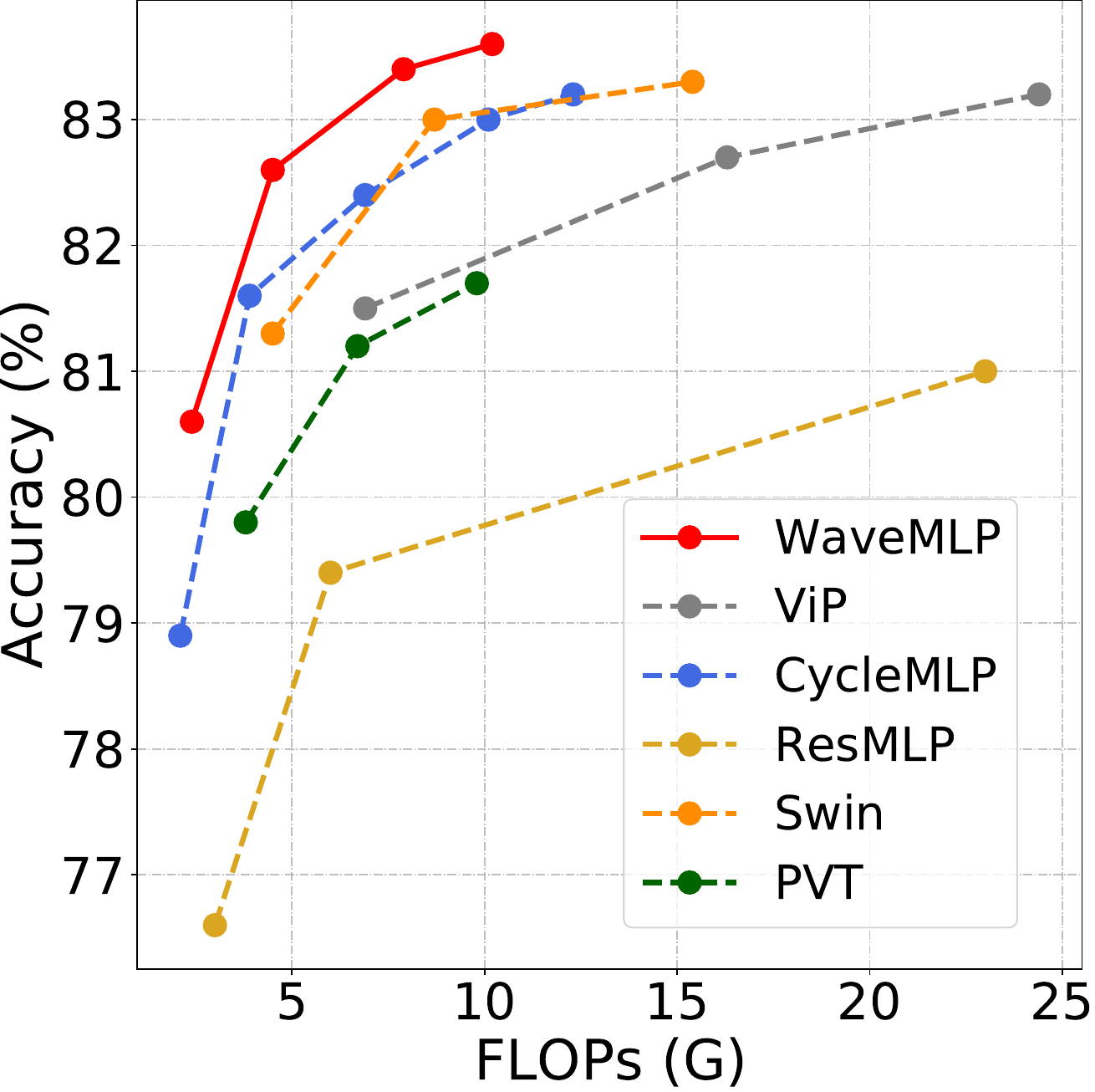}	
		\caption{Accuracy \wrt FLOPs.}
		\label{fig-accflops}
	\end{subfigure}
	\begin{subfigure}{0.48\linewidth}
		\includegraphics[width=1.0\linewidth]{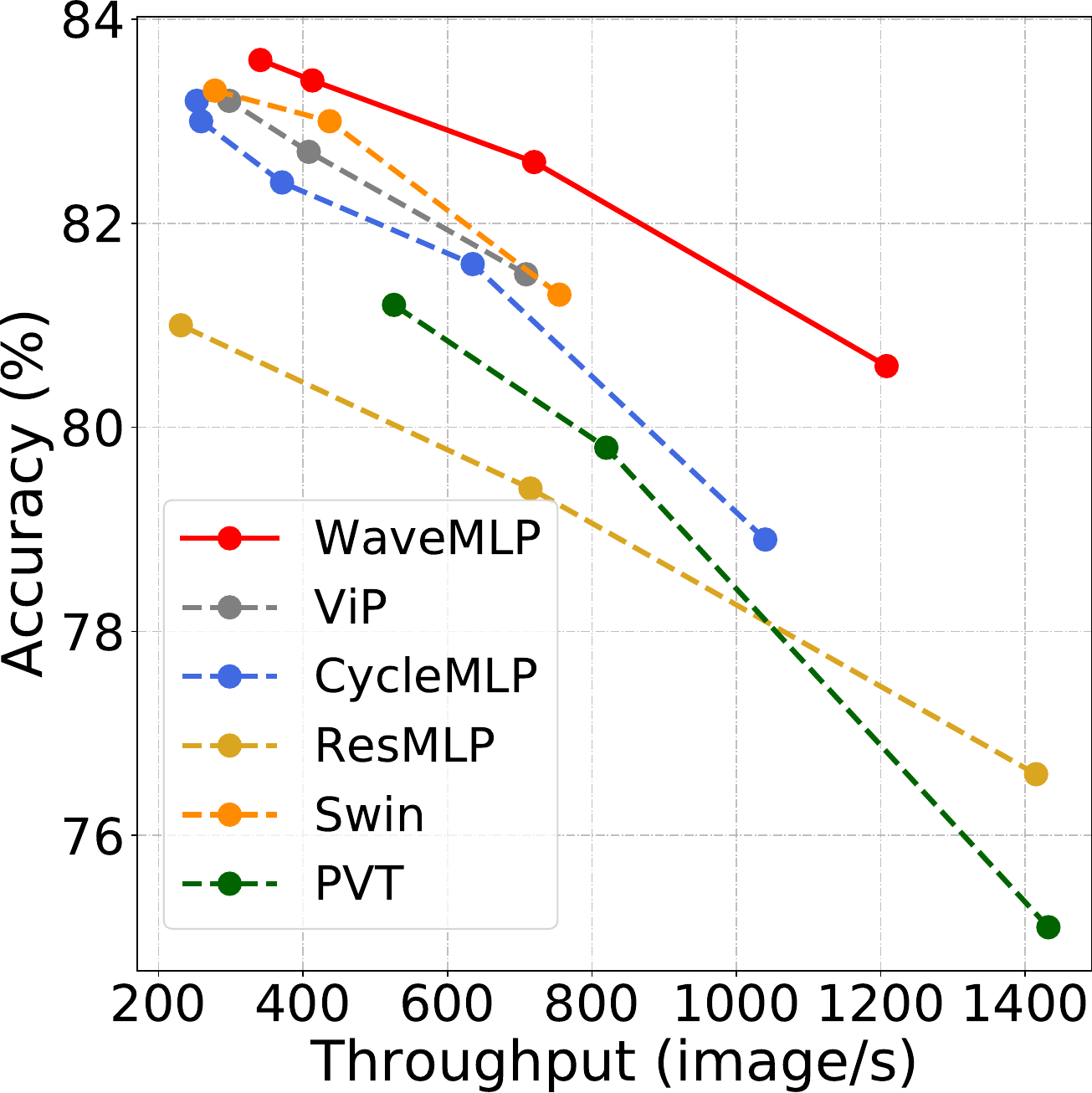}
		\caption{Accuracy \wrt throughput.}
		\label{fig-accth}
	\end{subfigure}
	\caption{Performance comparison between the proposed Wave-MLP and existing architectures. Top-1 accuracies on ImageNet are reported.}
	\label{fig-acc}
	\vspace{-3mm}
\end{figure}

Taking a sequence of image patches (tokens) as input, MLP-like models~\cite{tolstikhin2021mlp,touvron2021resmlp} mainly contain two separable blocks, \ie, channel-mixing MLP and token-mixing MLP, both composing of full-connected layers and activation functions. The channel-mixing MLP transforms feature of each token and the token-mixing MLP tries to aggregate information from different tokens.  By stacking these two types of MLP block alternatively, the simple MLP architecture could have sufficient capacity to extract features and achieve good performance on vision tasks.

However, the performance of MLP architecture is still inferior to that of SOTA Transformer and CNN architectures. We point out that one of the  bottlenecks for vision MLP lies in its manner of aggregating different tokens, \ie, mixing different tokens  with fixed weights of fully-connected layers. Recall that Transformer~\cite{vaswani2017attention, dosovitskiy2020image} aggregates tokens with weights dynamically adjusted by the attention mechanism. The inner products between  different tokens are calculated and tokens with higher  similarities  tend to have larger weights in the aggregation process of each other. However, the existing vision MLP models aggregate different tokens with fixed weights. 
The same weights are used for  tokens from different input images, neglecting differences in  semantic information of various  tokens, which may not aggregate  tokens well for all the input images.

Different from Transformer that delicately designs the attention mechanism, we aim to improve the representation way of tokens for dynamically aggregating them according to their semantic contents.  Actually, in quantum mechanics, an entity~(\eg, electron, photon) is usually represented by a wave function (\eg, de Broglie wave) containing both amplitude and phase~\cite{griffiths2018introduction, arndt1999wave,heller1994scattering}. The amplitude part measures the maximum intensity of a wave and the phase part  modulates the intensity  by indicating the location of a point in the wave period. Inspired by the quantum mechanics, we describe each token as a wave to realize the dynamic  aggregation procedure of tokens.

\begin{figure}[t] 
 	\centering
 	\small
 	\includegraphics[width=0.9\linewidth]{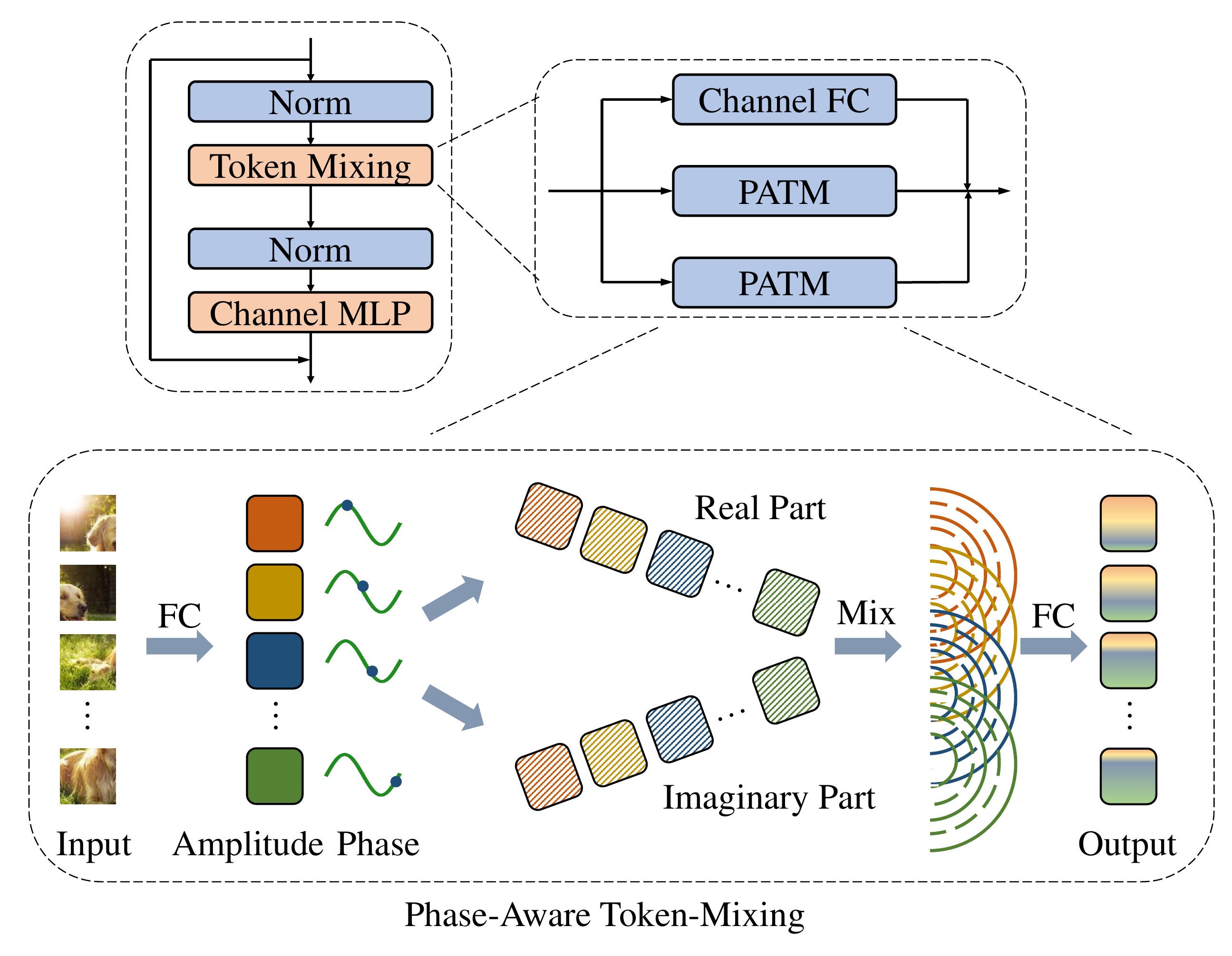}		
 	\caption{The diagram of a block in the Wave-MLP architecture.} 
 	\label{fig-pipeline}
 	\vspace{-6mm}
 \end{figure}

In this paper, we present a novel vision MLP architecture (dubbed as Wave-MLP), which takes each token as a wave with both amplitude and phase. The amplitude is the real-value feature representing the content of each token, while the phase term is a unit complex value  modulating the relationship between  tokens and fixed weights in MLP. The phase difference between these wave-like tokens affects their aggregated output and tokens with close phases tend to enhance each other.
Considering that tokens from different input images contain diverse semantic contents,
 we use a simple module to dynamically estimate the phase for each  token.  With  tokens equipped with amplitude and phase information, we introduce a phase-aware token mixing module~(PATM in Figure~\ref{fig-pipeline}) to aggregate these tokens. The whole Wave-MLP architecture is constructed by stacking the PATM module and channel-mixing MLP, alternately.

The proposed Wave-MLP architecture shows a large superiority to the existing architectures (shown in Figure~\ref{fig-acc}). For example, the proposed Wave-MLP-S model achieves 82.6\% top-1 accuracy on ImageNet with 4.5G FLOPs, which significantly surpasses Swin-T~\cite{liu2021Swin} with 81.3\% accuracy and 4.5G FLOPs. Besides, Wave-MLP also achieves strong performance on the dense prediction tasks such as  object detection and semantic segmentation.

The paper is organized as follows: Section~\ref{sec-related} briefly reviews the existing works about designing model architectures, and Section~\ref{sec-method} discusses the proposed  Wave-MLP architecture detailedly. In Section~\ref{sec-exp}, we empirically investigate the method's effectiveness on multiple vision tasks and make conclusions in Section~\ref{sec-con}.

\section{Related Work}
\label{sec-related}

\noindent\textbf{CNN-based Architectures.} Convolutional neural networks (CNNs) have been the mainstream in computer vision for a long time. The prototype of CNN model is presented in \cite{lecun1998gradient} for the document recognition task, where convolution is the core operation. Beginning with great success of AlexNet~\cite{krizhevsky2012imagenet}  in ILSVRC~2012, various architectures such as GoogleNet~\cite{szegedy2015going}, VGGNet~\cite{simonyan2014very}, ResNet~\cite{he2016deep}, RegNet~\cite{radosavovic2020designing} are developed. Though the model architectures become more complex for pursuing high performance, the core operations have always the convolution and its variants. The occurrence of new computing paradigm such as vision Transformer~\cite{vaswani2017attention}, vision MLP~\cite{tolstikhin2021mlp} bring new blood to the area of architecture design in computer vision.

\noindent\textbf{Transformer-based Architectures.}   Transformer~\cite{vaswani2017attention} is originally proposed for the natural language processing (NLP) tasks such as  language modeling and machine translation. Dosovitskiy~\etal~\cite{dosovitskiy2020image} introduce it to computer vision and achieve excellent performance on image classification tasks especially when training data are extremely sufficient. Touvron~\etal~\cite{touvron2021training} refine the training recipe and present a teacher-student strategy specific to transformers, which produce competitive transformer models trained on ImageNet from scratch. Then many works  explore the architecture design of vision transformers~\cite{wu2021rethinking,han2021transformer,tang2021patch,wu2020visual,tang2021augmented,chen2021crossvit,guo2021cmt,han2022pyramidtnt}. For example, Han~\etal~\cite{han2021transformer} present a nested transformer architecture to capture global and local information simultaneously. To be compatible with the dense prediction task such as object detection and semantic segmentation,    hierarchical architectures are adopted in \cite{wang2021pyramid,dosovitskiy2020image,heo2021rethinking}, which splits the whole architecture into multiple stages and reduce the spatial resolution stage-wisely. Swin Transformer~\cite{liu2021Swin} extract representation with shifted windows and limit the self-attention in local regions. Compared with  the self-attention in \cite{dosovitskiy2020image} connecting all the tokens in a layer, the shifted window operation is more efficient.

\noindent\textbf{MLP-based Architectures.} Recently, MLP-like architectures composing of fully connected layers and non-linear activation functions have been paid much attention~\cite{tolstikhin2021mlp,chen2021cyclemlp,guo2021hire,li2022brain}. Though they have more simple architectures and introduce less inductive bias, their performances are still comparable with SOTA  models.  The MLP-Mixer model~\cite{tolstikhin2021mlp} uses two type of MLP layers, \ie, channel-mixing MLP and token-mixing MLPs. The channel-MLP extract features for each tokens while the token-mixing MLPs capture the spatial information. Touvron~\etal~\cite{touvron2021resmlp} present a similar architecture and replace the Layer Normalization~\cite{ba2016layer} with the simpler affine transformation. Liu~\etal~\cite{liu2021pay} empirically validate that MLP architectures with gating can achieve similar performance with Transformers in both language and vision tasks. To preserve the positional information of input images, Hou~\etal~\cite{hou2021vision} keep the 2D shape of the input image and extract features by permuting them along width and height, respectively. Based on MLP-Mixer, Yu~\etal~\cite{yu2021s} replace the token-mixing MLP with a spatial shift operation for capturing the local spatial information, which is also computationally  efficient. Currently, Lian~\etal~\cite{lian2021mlp} propose to shift tokens along two orthogonal directions to obtain an axial receptive field. Chen~\etal~\cite{chen2021cyclemlp} propose a cycle fully-connected layer, which mixes information along the spatial and channel dimensions simultaneously and can cope with variable input image scales. Different from them, we explore how to represent the tokens in vision MLP and take each token as a wave with both amplitude and phase. Empirically, we find that our Wave-MLP architecture achieves a better trade-off between accuracy and computational cost compared with the existing architectures.

\section{Method}
\label{sec-method}

In this section, we discuss the proposed Wave-MLP models detailedly.
After introducing the vision MLP architecture briefly, we present the phase-aware token mixing module~(PATM), which represents each token as a wave and aggregate them by considering amplitude and phase simultaneously. At last, we describe the blocks in Wave-MLP and architecture variants with different computational costs.

\subsection{Preliminaries}

A MLP-like model is a neural architecture mainly composed of full-connected layers and non-linear activation functions. For the vision MLP, it first splits an image into multiple patches (also referred to as tokens) and  then extract their features with two components, \ie, channel-FC and token-FC described as following.

Denote the intermediate feature containing $n$ tokens as $Z=[\z_1,\z_2,\cdots,\z_n]$, where each token $\z_j$ is a $d$-dimension vector. The channel-FC is formulated as:
\begin{equation}
\label{eq-cfc}
\text{Channel-FC}(\z_j,W^c)=W^c \z_j, j=1,2,\cdots n,
\end{equation}
where $W^c$ is the  weight with learnable parameters. The channel-FC operates on each token independently to extract their features. To enhance the transformation ability, multiple channel-FC layers are usually stacked together with the non-linear activation function, which constructs a channel-mixing MLP. 

To aggregate information from different tokens, the token-FC operation is required, \ie,
\begin{equation}
\label{eq-tfc}
\text{Token}\text{-FC}(Z, W^t)_j=  \sum_{k} W^t_{jk} \odot \z_k, j=1,2,\cdots n,
\end{equation} 
where $W^t$ is the token-mixing weight, $\odot$ denotes element-wise multiplication and the subscript $j$ indicates the $j$-th output token. The token-FC operation tries to capture the spatial information by mixing features from different tokens. In the existing MLP-like models such as MLP-Mixer~\cite{tolstikhin2021mlp}, ResMLP~\cite{touvron2021resmlp}, a token-mixing MLP is also constructed by stacking the token-FC layers and activation functions.  Such a simple token-mixing MLP with fixed weights neglects the varying semantic contents of tokens from different input images, which is a bottleneck restricting the representation ability of MLP-like architectures.

\begin{figure}[t] 
	\centering
	\small
	\includegraphics[width=0.8\linewidth]{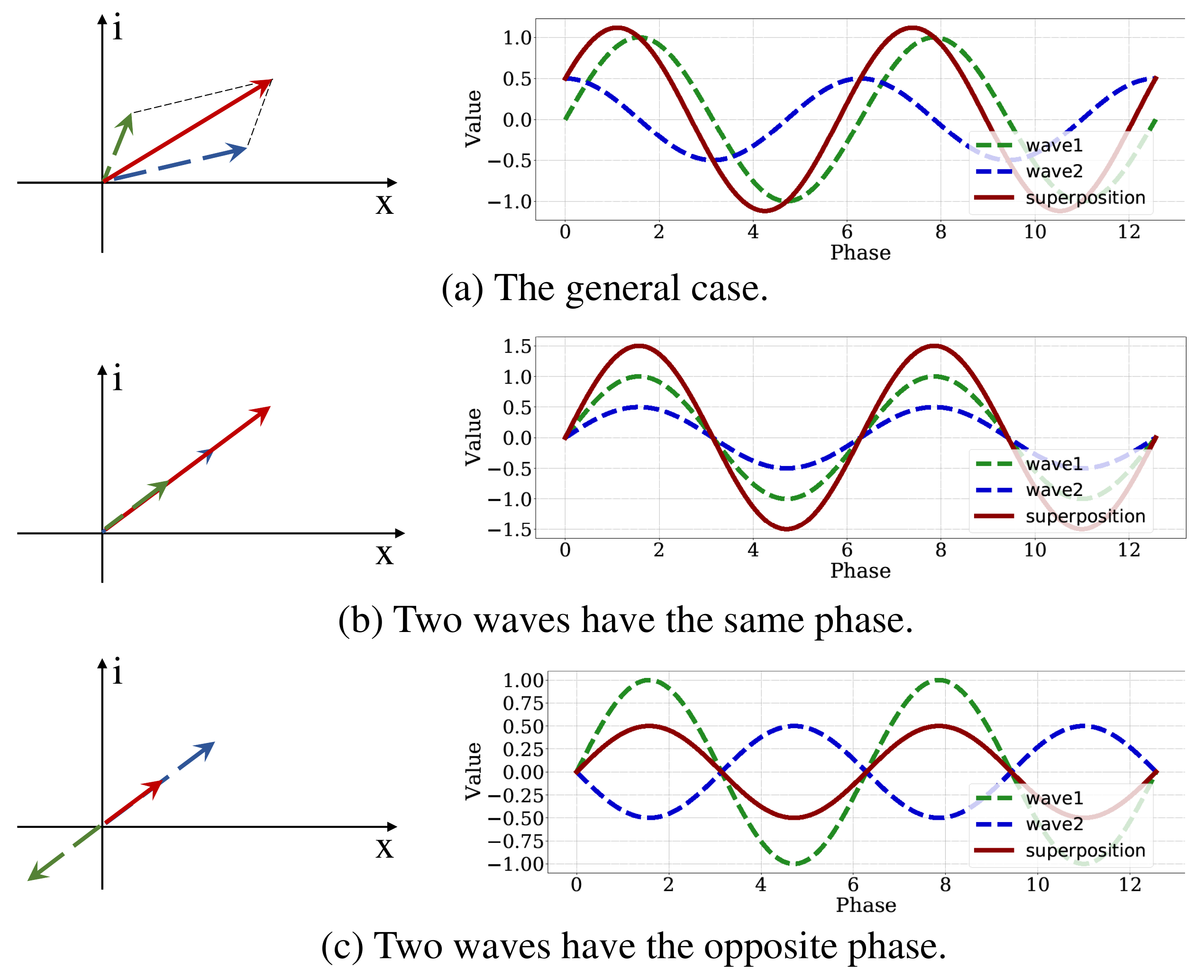}
	\caption{The interaction between two waves with different phase. The left is the superposition of two waves in the complex-value domain, while the right shows how their projections along the real axis varies \wrt the phase. The dashed lines denote two waves with different initial phase, and the solid line is their superposed wave. }
	\label{fig-wave}
	\vspace{-4mm}
\end{figure}

\subsection{Phase-Aware Token Mixing}
\label{sec-patm}

To dynamically modulate the relationship between  tokens and fixed weights in MLP for aggregating tokens more properly, we take each token as a wave with both amplitude and phase. We firstly discuss the wave-like representation of a token and then present the phase-aware token mixing module (PATM)  for aggregating tokens.

\noindent\textbf{Wave-like representation.} In Wave-MLP, a token is represented as a wave $\tilde \z_j$ with both amplitude and phase information, \ie, 
\begin{equation} 
	\label{eq-token}
	\tilde \z_j = |\z_j| \odot e^{i\btheta_j}, j=1,2,\cdots,n,
\end{equation} 
where $i$ is the imaginary unit satisfying $i^2=-1$. $|\cdot|$ denotes the absolute value operation and $\odot$ is element-wise  multiplication.   The amplitude $|\z_j|$ is a real-value feature representing the content of each token.
$e^{i\btheta_j}$ is a periodic function whose elements always have the unit norm. $\btheta_j$ indicates the phase, which is the current location of  token within a  wave period. With both amplitude and phase, each token $\tilde \z_j $ is represented in the complex-value domain.

When aggregating different tokens, the phase term $\btheta_j$ modulates their superposition modes. Supposing $\tilde \z_r=\tilde \z_1 + \tilde \z_2$  is the aggregated results of wave-like token $\tilde \z_1$, $\tilde \z_2$ \footnote{Without affecting the conclusion, the aggregating weights are set to 1 for simplicity.}, its amplitude $|\z_r|$ and phase $\btheta_r$ can be calculated  as following:
\begin{equation}
	\label{eq-amp2}
	\small
	|\z_r| = \sqrt{|\z_i|^2 + |\z_j|^2 + 2 |\z_i|\odot |\z_j|\odot \cos(\btheta_j-\btheta_i)}, 
\end{equation}
\begin{equation}
	\label{eq-phs2}
	\small
	\begin{aligned}	
		\btheta_r= \btheta_i + {\rm atan2}(&|\z_j|\odot \sin(\btheta_j-\btheta_i),\\
		&|\z_i|+|\z_j|\odot \cos(\btheta_j-\btheta_i)),
	\end{aligned}
\end{equation} 
where ${\rm atan2}(x,y)$ is the two-argument arctangent function. As shown in the above equations, the phase difference $ |\btheta_j-\btheta_i|$ between two tokens  has a large impact on the amplitude of aggregated result $\z_r$. An intuitive diagram  is shown in Figure~\ref{fig-wave}. The left is the superposition of two waves in the complex-value domain, while the right shows how their projections along the real axis varies \wrt the phase. When two tokens have the same phase ($\btheta_j=\btheta_i+2\bm \pi*m, m\in [0, \pm 2, \pm 4, \cdots]$), they will be enhanced by each other, \ie, $|\z_r| = |\z_i| +|\z_j|$ (Figure~\ref{fig-wave} (b)). For the opposite phase ($\btheta_j=\btheta_i+\bm \pi*m, m\in [\pm 1, \pm 3, \cdots]$), the resultant wave will be weakened ($|\z_r| = ||\z_i| -|\z_j||$).  In other cases, their interaction is more complex but whether they will be enhanced or weakened also depends on the phase difference (Figure~\ref{fig-wave} (a)). Note that the classical representation strategy with only real-value feature is a special case of Eq~\ref{eq-token},  whose phase $\btheta_j$ is only the integer multiple of $\bm \pi$.

\noindent\textbf{Amplitude.} To get the wave-like tokens in Eq.~\ref{eq-token}, both amplitude and phase information are required. The amplitude $|\z_i|$ is similar to the real-value feature in the traditional model, expect for an absolute operation. Actually, the element-wisely absolute operation can be absorbed into the phase term, \ie, $|z_{j,t}|e^{i\theta_{j,t}}=z_{j,t}e^{i\theta_{j,t}}$ if $z_{j,t}>0$, and $|z_{j,t}|e^{i\theta_{j,t}}=z_{j,t}e^{i(\theta_{j,t}+\pi)}$ otherwise, where $z_{j,t}$ and $\theta_{j,t}$ denote the $t$-th element in $\z_j$ and $\btheta_j$. Thus we remove the absolute operation in practical implementation for simplicity. Denoting $X=[\x_1,\x_2,\dots,\x_n]$ as the input of a block, we get the token's amplitude $\z_j$  by a plain channel-FC operation, \ie,
 \begin{equation}
 	\z_j=\text{Channel-FC}(\x_j,W^c), j=1,2,\cdots,n.
 \end{equation}

\noindent\textbf{Phase.}
Recalling that the phase indicates the current location of token in a period of wave,  we discuss different strategy to generate phases as following. The simplest strategy (`static phase') is to represent the phase $\btheta_j$ of each tokens with fixed parameters, which can be learned in the training process. Though the static phase can distinguish different tokens, it  neglects the diversity of different input images either.

To capture the particular attributes for each input respectively, we use an estimation module $\Theta$ to generate the phase information according to input features $\x_j$, \ie,  $\bm \theta_j=\Theta(\x_j,W^\theta)$, where $W^\theta$ denotes the learnable parameters.
Considering that simplicity is an important characteristic of MLP-like architectures, complex operations are undesirable. Thus we also adopt the simple channel-FC in Eq.~\ref{eq-cfc} as the phase estimation module. The estimation module can also be constructed with other formulations, whose impact on the model performance is  empirically investigated  in Table~\ref{tab-form} of Section~\ref{sec-abl}.

\noindent\textbf{Token aggregation.} In Eq.~\ref{eq-token}, the wave-like tokens are represented in the complex domain. To embed it in a general MLP-like architecture, we unfold it with Euler's formula and represent it with real part and imaginary part, \ie,
\begin{equation}
	\label{eq-unfold}
	\tilde \z_j = |\z_j|\odot \cos\bm\theta_j+i|\z_j|\odot \sin\bm\theta_j, j=1,2,\cdots,n. 
\end{equation}
In the above equation, a complex-value token is represented as two real-value vectors, indicating the  real and imaginary parts, respectively. Different tokens $\tilde \z_j$ are then aggregated with the token-FC operation~(Eq.~\ref{eq-tfc}), \ie,
\begin{equation} 
	\label{eq-agg}
	\tilde \o_j= \text{Token-FC}(\tilde Z,  W^t)_j, j=1,2,\cdots, n,
\end{equation}
where $\tilde Z=[\tilde \z_1,\tilde \z_2, \cdots,\tilde \z_n]$ denotes all the wave-like tokens in a layer.
 In Eq.~\ref{eq-agg}, different tokens interact with each other considering both the amplitude and phase information. The output $\tilde \o_j$ is the complex-value representation of the aggregated feature. Following the common quantum measurement methods~\cite{braginsky1995quantum,jacobs2006straightforward} which project a quantum state with complex-value representation to the real-value observable, we  get the real-value output $\o_j$ by summing the real and imaginary part of $\tilde \o_j$ with weights. Combined with Eq.~\ref{eq-agg}, the output $\o_j$ can be obtained as:
\begin{equation} 
	\label{eq-spatial}
	\begin{aligned}
\o_j = &\sum_{k}W^t_{jk}\z_k\odot \cos\bm\theta_k+W^i_{jk}\z_k\odot \sin\bm\theta_k, \\
&j=1,2,\cdots, n, 
	\end{aligned}
\end{equation}
where $W^t$, $W^i$ are both learnable weights.
In the above equation, the phase $\bm \theta_k$ adjusts dynamically according to the semantic content of input data. Besides the fixed weights, the phases also modulate the aggregating process of different tokens.

In vision MLP, we construct a phase-aware token mixing module (PATM) to conduct the above token aggregating procedure, which is shown in Figure~\ref{fig-pipeline}. Given the input feature $\x_j$, the amplitude $\z_j$ and phase $\bm \theta_j$ are generated with the channel-FC and phase estimation module, respectively. Then the wave-like token $\tilde \z_j$ is unfolded with Eq.~\ref{eq-unfold} and aggregated to get output feature $\o_j$ (Eq.~\ref{eq-spatial}). The final module output  is obtained by transforming $\o_j$ with another channel-FC to enhance the representation capacity.


\subsection{Wave-MLP Block}

A basic unit in the proposed Wave-MLP mainly contains two blocks, channel-mixing MLP and phase-aware token-mixing block~(Figure~\ref{fig-pipeline}). The channel-mixing MLP is stack by two channel-FC layers~(Eq.~\ref{eq-cfc}) and non-linear activation functions, which extracts features for each token. The token-mixing block composes of the proposed PATM modules, aggregating different tokens by considering both amplitude and phase information.

To be more compatible with computer vision tasks, we preserve the 2D spatial shape of input image by using feature maps with shape $H\times W\times C$, which $H$, $W$, $C$ are the height, width and channel's number, respectively. This is a successful practice widely used in recent vision transformer architectures (\eg, PVT~\cite{wang2021pyramid}, Swin-Transformer~\cite{liu2021Swin}). There are two parallel PATM modules, which aggregate spatial information along high and width dimensions, respectively. Similar to \cite{hou2021vision,chen2021cyclemlp}, different branches are
summed with a re-weighting module. In the traditional MLP-Mixer~\cite{tolstikhin2021mlp}, each  token-FC layer connects all tokens together, whose dimension depends on specific input size. Thus it is not compatible with the dense prediction tasks (\eg, object detection and semantic segmentation) with varying sizes of input images. To address this issue,  we use a simple strategy that restricts the FC layers only connect tokens within a local window. The empirical investigation of the window size is shown in Table~\ref{tab-size} of Section~\ref{sec-abl}. Besides the PATM modules, another channel-FC connecting the input and output directly is also used to preserve the original information. The final output of the block is the summation of these three branches.

The whole model is constructed by stacking phase-aware token-mixing blocks, channel-mixing MLPs,  and normalization layers, alternately. 
 To produce hierarchical features, we split the architecture into 4 stages, which reduces the size of feature maps and increases the number of channels stage-wisely. By varying the width and depth of model, we develop 4 models with different parameters and computational costs, denoted as Wave-MLP-T, Wave-MLP-S, Wave-MLP-M, Wave-MLP-B, sequentially. The detailed configures of these models can be found in the supplemental material.

\begin{table}
	\centering
	\small 
	\caption{Comparison of the proposed Wave-MLP architecture with existing vision  MLP models on ImageNet.} 
	\label{tab-mlp}

	\setlength\tabcolsep{3.4pt}
	\begin{tabular}{l|ccc|c}
	
		\toprule[1.5pt]
	
		\multirow{2}{*}{Model} &  \multirow{2}{*}{Params.} & \multirow{2}{*}{FLOPs} & Throughput &  Top-1 \\ 
		&&&(image / s)&acc. (\%) \\ \hline
	
		EAMLP-14~\cite{guo2021beyond}          & 30M  & -    &771 &78.9 \\
		EAMLP-19~\cite{guo2021beyond}                    & 55M  & -    &464& 79.4 \\ \hline
		
		Mixer-B/16~\cite{tolstikhin2021mlp}            & 59M  & 12.7G&- & 76.4  \\ 
		\hline
		
		ResMLP-S12~\cite{touvron2021resmlp}                & 15M  & 3.0G  & 1415 &76.6 \\
		ResMLP-S24~\cite{touvron2021resmlp}                 & 30M  & 6.0G  &715& 79.4 \\
		ResMLP-B24~\cite{touvron2021resmlp}                           & 116M & 23.0G &231& 81.0 \\ \hline

		gMLP-S~\cite{liu2021pay}                                & 20M  & 4.5G  &-& 79.6 \\
		gMLP-B~\cite{liu2021pay}                                & 73M  & 15.8G &-& 81.6 \\ \hline
		
		S$^2$-MLP-wide~\cite{yu2021s}           & 71M  & 14.0G &-& 80.0 \\
		S$^2$-MLP-deep~\cite{yu2021s}                      & 51M  & 10.5G &-& 80.7 \\ \hline

		ViP-Small/7~\cite{hou2021vision}                  & 25M   & 6.9G  &719& 81.5 \\
		ViP-Medium/7~\cite{hou2021vision}                  & 55M   & 16.3G &418& 82.7 \\
		ViP-Large/7~\cite{hou2021vision}                  & 88M   & 24.4G &298& 83.2\\ \hline

		AS-MLP-T~\cite{lian2021mlp}              & 28M   & 4.4G  &862& 81.3 \\
		AS-MLP-S~\cite{lian2021mlp}                & 50M   & 8.5G  &473&83.1 \\
		AS-MLP-B~\cite{lian2021mlp}                   & 88M   & 15.2G  &308& 83.3\\  \hline
		
		CycleMLP-B1~\cite{chen2021cyclemlp}    & 15M   & 2.1G  &1040& 78.9 \\
		CycleMLP-B2~\cite{chen2021cyclemlp}           & 27M   & 3.9G  &635& 81.6 \\
		CycleMLP-B3~\cite{chen2021cyclemlp}       & 38M   & 6.9G  &371& 82.4 \\
		CycleMLP-B4~\cite{chen2021cyclemlp}        & 52M   & 10.1G &259& 83.0 \\
		CycleMLP-B5~\cite{chen2021cyclemlp}         & 76M   & 12.3G &253& 83.2 \\  \hline
	Wave-MLP-T*    (ours)                     & 15M   & 2.1G  &1257& \textbf{80.1} \\
		Wave-MLP-T (ours)                       & 17M   & 2.4G  &1208& \textbf{80.6} \\	
		Wave-MLP-S  (ours)                                & 30M   & 4.5G  &720 &\textbf{82.6} \\
		Wave-MLP-M      (ours)                              & 44M   & 7.9G  &413& \textbf{83.4}\\
		Wave-MLP-B   (ours)                                 & 63M   & 10.2G &341&\textbf{83.6} \\
		\bottomrule[1.5pt]
	\end{tabular}
\vspace{-2mm}
\end{table}

\begin{table}
	\centering
	\small 
	\caption{Comparison of the proposed Wave-MLP architecture with SOTA models on ImageNet.} 
	\label{tab-sota}

	\setlength\tabcolsep{2pt}
\begin{tabular}{l | c |c c c|c}
	
	\toprule[1.5pt]

	\multirow{2}{*}{Model} & \multirow{2}{*}{Family}  & \multirow{2}{*}{Params.} & \multirow{2}{*}{FLOPs} & Throughput &  Top-1 \\ 
	&&&&(image / s)&acc. (\%)\\ \hline

	ResNet18~\cite{he2016deep}                   & CNN    & 12M & 1.8G  &-& 69.8 \\
	ResNet50~\cite{he2016deep}                   &  CNN   & 26M & 4.1G  &- &78.5 \\
	ResNet101~\cite{he2016deep}                  &  CNN  & 45M & 7.9G &-& 79.8 \\
	RegNetY-4G~\cite{radosavovic2020designing}   &  CNN   & 21M & 4.0G  &1157& 80.0 \\
	RegNetY-8G~\cite{radosavovic2020designing}   &  CNN  &  39M & 8.0G &592& 81.7 \\
	RegNetY-16G~\cite{radosavovic2020designing}   &  CNN  &  84M & 16.0G &335& 82.9 \\
	\hline	
	GFNet-H-S~\cite{rao2021global}               & FFT   &  32M & 4.5G  & -&81.5 \\
		GFNet-H-B~\cite{rao2021global}               & FFT   &  54M & 8.4G  & -&82.9 \\
		BoT-S1-50~\cite{srinivas2021bottleneck}      & Hybrid&  21M & 4.3G  &-& 79.1 \\
	BoT-S1-59~\cite{srinivas2021bottleneck}      & Hybrid&  34M & 7.3G &-& 81.7  \\ \hline

	DeiT-S~\cite{touvron2021training}            & Trans  & 22M & 4.6G  & 940&79.8 \\
	DeiT-B~\cite{touvron2021training}            & Trans &  86M & 17.5G &  292&81.8 \\
		PVT-Small~\cite{wang2021pyramid}                 & Trans &  25M & 3.8G  &820& 79.8 \\
		PVT-Medium~\cite{wang2021pyramid}                 & Trans &  44M & 6.7G &526& 81.2 \\
		PVT-Large~\cite{wang2021pyramid}                 & Trans &  61M & 9.8G  &367& 81.7 \\
	T2T-ViT-14~\cite{yuan2021tokens} &Trans &22M&5.2G&764&81.5\\
	T2T-ViT-19~\cite{yuan2021tokens} &Trans &39M&8.9G&464&81.9\\
	T2T-ViT-24~\cite{yuan2021tokens} &Trans&64M&14.1G&312&82.3\\
	TNT-S~\cite{han2021transformer} & Trans  & 24M & 5.2G  &428& 81.5 \\
	TNT-B~\cite{han2021transformer}& Trans &  66M & 14.1G  &246& 82.9 \\
		iRPE-K~\cite{wu2021rethinking} &Trans & 87M&17.7G&- &82.4\\
	iRPE-QKV~\cite{wu2021rethinking}            & Trans  & 22M & 4.9G  &-& 81.4 \\

	GLiT-Small~\cite{chen2021glit}& Trans&25M&4.4G&-&80.5\\

	GLiT-Base~\cite{chen2021glit}& Trans&96M&17.0G&-&82.3\\
		Swin-T~\cite{liu2021Swin}                    & Trans &  29M & 4.5G  &755& 81.3 \\
		Swin-S~\cite{liu2021Swin}                    & Trans & 50M & 8.7G  & 437&83.0 \\

	Swin-B~\cite{liu2021Swin}                    & Trans &  88M & 15.4G &278 &83.5 \\
\hline
	Wave-MLP-T*                                  & MLP     & 15M & 2.1G  &1257&\textbf{80.1} \\  
Wave-MLP-T                      & MLP & 17M   & 2.4G  &1208& \textbf{80.6} \\	
	Wave-MLP-S                             & MLP    & 30M   & 4.5G  &720 &\textbf{82.6} \\
		Wave-MLP-M                            & MLP     & 44M   & 7.9G  &413& \textbf{83.4}\\
	Wave-MLP-B                               & MLP     & 63M   & 10.2G &341&\textbf{83.6} \\

	\bottomrule[1.5pt]
	
\end{tabular}
\vspace{-3mm}
\end{table}

\begin{table*}
	\centering
	\small 
	\caption{Results of object detection and instance segmentation on COCO val2017.} 
	\label{tab-coco}
	\setlength\tabcolsep{3.4pt}
\begin{tabular}{l|c |lcc|lcc| c|lcc|lcc}

	\toprule[1.5pt]
	\multirow{2}{*}{Backbone} &\multicolumn{7}{c|}{RetinaNet 1$\times$} &\multicolumn{7}{c}{Mask R-CNN 1$\times$} \\
	\cline{2-15} 
	& Params. / FLOPs & AP &AP$_{50}$ &AP$_{75}$ &AP$_S$ &AP$_M$ &AP$_L$ & Params. / FLOPs& AP$^{\rm b}$ &AP$_{50}^{\rm b}$ &AP$_{75}^{\rm b}$  &AP$^{\rm m}$ &AP$_{50}^{\rm m}$ &AP$_{75}^{\rm m}$\\ \hline
	
	ResNet18~\cite{he2016deep}                & 21.3M / 188.7G & 31.8 & 49.6 & 33.6 & 16.3 & 34.3 & 43.2 & 31.2M / 207.3G & 34.0 & 54.0 & 36.7 & 31.2 & 51.0 & 32.7 \\
	PVT-Tiny~\cite{wang2021pyramid}           & 23.0M / 189.5G & 36.7 & 56.9 & 38.9 & 22.6 & 38.8 & 50.0 & 32.9M / 208.1G & 36.7 & 59.2 & 39.3 & 35.1 & 56.7 & 37.3 \\
	CycleMLP-B1~\cite{chen2021cyclemlp}                           & 24.9M / 195.0G& 38.6 & 59.1 & 40.8 & 21.9 & 41.8 & 50.7 & 34.8M / 213.6G& 39.4 & 61.4 & 43.0 & 36.8 & 58.6 & 39.1 \\
	Wave-MLP-T                           & 25.3M / 196.3G &\textbf{40.4}&61.0&43.4&24.9&43.7&51.7&35.2M / 214.6G&\textbf{41.5}&63.7&45.4&38.2&60.9&40.7\\
	\hline
	
	ResNet50~\cite{he2016deep}                & 37.7M / 239.3G & 36.3 & 55.3 & 38.6 & 19.3 & 40.0 & 48.8 & 44.2M / 260.1G & 38.0 & 58.6 & 41.4 & 34.4 & 55.1 & 36.7 \\
	
	Swin-T~\cite{liu2021Swin} & 38.5M / 244.8G & 41.5 &62.1&44.2& 25.1 & 44.9 & \textbf{55.5} & 47.8M / 264.0G & 42.2 & 64.6 & 46.2 & 39.1 & 61.6 & 42.0 \\ 
	PVT-Small~\cite{wang2021pyramid}          & 34.2M /226.5G & 40.4 & 61.3 & 43.0 & 25.0 & 42.9 & 55.7 & 44.1M / 245.1G & 40.4 & 62.9 & 43.8 & 37.8 & 60.1 & 40.3 \\
	CycleMLP-B2~\cite{chen2021cyclemlp}                           & 36.5M / 230.9G& 40.9 & 61.8 & 43.4 & 23.4 & 44.7 & 53.4 & 46.5M /249.5G & 41.7 & 63.6 & 45.8 & 38.2 & 60.4 & 41.0 \\
	
	Wave-MLP-S                           & 37.1M / 231.3G &\textbf{43.4} &64.4&46.5&26.6&47.1&57.1 & 47.0M /250.3G &\textbf{44.0} &65.8&48.2&40.0&63.1&42.9 \\
	\hline
	
	ResNet101~\cite{he2016deep}               & 56.7M / 315.4G & 38.5 & 57.8 & 41.2 & 21.4 & 42.6 & 51.1 & 63.2M / 336.4G & 40.4 & 61.1 & 44.2 & 36.4 & 57.7 & 38.8 \\

	Swin-S~\cite{liu2021Swin}&59.8M / 334.8G & 44.5 &65.7&47.5& 27.4 & 48.0 & 59.9 & 69.1M / 353.8G & 44.8 & 66.6 & 48.9 & 40.9 & 63.4 & 44.2 \\
	PVT-Medium~\cite{wang2021pyramid}         & 53.9M / 283.1G & 41.9 & 63.1 & 44.3 & 25.0 & 44.9 & 57.6 & 63.9M / 301.7G & 42.0 & 64.4 & 45.6 & 39.0 & 61.6 & 42.1 \\

	CycleMLP-B3~\cite{chen2021cyclemlp}                           & 48.1M / 291.3G & 42.5 & 63.2 & 45.3 & 25.2 & 45.5 & 56.2 & 58.0M / 309.9G & 43.4 & 65.0 & 47.7 & 39.5 & 62.0 & 42.4 \\ 
	
	Wave-MLP-M                           & 49.4M / 291.3G &\textbf{44.8} &65.8&47.8&28.0&48.2&59.1&59.6M / 311.5G&\textbf{45.3}&67.0&49.5&41.0&64.1&44.1 \\ 
	
	\hline
	
	PVT-Large~\cite{wang2021pyramid}          & 71.1M / 345.7G & 42.6 & 63.7 & 45.4 & 25.8 & 46.0 & 58.4 & 81.0M / 364.3G& 42.9 & 65.0 & 46.6 & 39.5 & 61.9 & 42.5 \\
	CycleMLP-B4~\cite{chen2021cyclemlp}                           & 61.5M / 356.6G & 43.2 & 63.9 & 46.2 & 26.6 & 46.5 & 57.4 & 71.5M / 375.2G & 44.1 & 65.7 & 48.1 & 40.2 & 62.7 & 43.5 \\
		CycleMLP-B5~\cite{chen2021cyclemlp}                           & 85.9M / 402.2G & 42.7 & 63.3 & 45.3 & 24.1 & 46.3 & 57.4 & 95.3M / 421.1G & 44.1 & 65.5 & 48.4 & 40.1 & 62.8 & 43.0 \\
	
	Wave-MLP-B &66.1M / 333.9G&\textbf{44.2}&65.1&47.1&27.1&47.8&58.9&75.1M / 353.2G&\textbf{45.7}&67.5&50.1&27.8&49.2&59.7\\

	\bottomrule[1.5pt]
\end{tabular}
\vspace{-4mm}
\end{table*}

\begin{table} 
	\centering
	\small 
	\caption{The semantic segmentation results of different backbones on the ADE20K validation set.   $^\dagger$ Results are from GFNet~\cite{rao2021global}.}
	\label{tab-ade}
\begin{tabular}{l|c|c|c}
	
	\toprule[1.5pt]
	\multirow{2}{*}{Backbone} & \multicolumn{3}{c}{Semantic FPN}\\
	\cline{2-4}
	& Params. & FLOPs&mIoU (\%)   \\ \hline

	ResNet18~\cite{he2016deep}                & 15.5M &127G& 32.9 \\
	PVT-Tiny~\cite{wang2021pyramid}           & 17.0M &123G& 35.7 \\
	CycleMLP-B1~\cite{chen2021cyclemlp}                   & 18.9M &130G& 39.5 \\
	Wave-MLP-T~(ours)                    & 19.3M &131G&  \textbf{41.2}\\
	\hline
	
	ResNet50~\cite{he2016deep}                & 28.5M &183G& 36.7 \\
	PVT-Small~\cite{wang2021pyramid}          & 28.2M &163G &39.8 \\
	Swin-S$^\dagger$~\cite{liu2021Swin}       & 31.9M &182G& 41.5 \\ 
	GFNet-H-Ti~\cite{rao2021global}           & 26.6M &126G& 41.0 \\
	CycleMLP-B2~\cite{chen2021cyclemlp}       & 30.6M &167G& 42.4 \\
	Wave-MLP-S~(ours)                   & 31.2M &168G& \textbf{44.4} \\
	\hline
	ResNet101~\cite{he2016deep}               & 47.5M &260G& 38.8\\

	PVT-Medium~\cite{wang2021pyramid}         & 48.0M &219G& 41.6 \\
	GFNet-H-S~\cite{rao2021global}          & 47.5M &179G& 42.5 \\

	Swin-B$^\dagger$~\cite{liu2021Swin}       & 53.2M &274G& 45.2 \\

	GFNet-H-B~\cite{rao2021global}           & 74.7M &261G& 44.8 \\
	CycleMLP-B3~\cite{chen2021cyclemlp}                    & 42.1M &229G& 44.5 \\
	CycleMLP-B4~\cite{chen2021cyclemlp}                  & 55.6M &296G& 45.1 \\
	CycleMLP-B5~\cite{chen2021cyclemlp}                    & 79.4M &343G& 45.6 \\
	Wave-MLP-M~(ours)            & 43.3M &231G& \textbf{46.8} \\
	\bottomrule[1.5pt]
\end{tabular}
\vspace{-2mm}
\end{table}

\section{Experiments}
\label{sec-exp}

In this section, we empirically investigate the proposed Wave-MLP architecture on multiple tasks, containing image classification,  object detection and semantic segmentation. Wave-MLP is firstly compared with the existing vision MLPs, vision Transformers and CNNs on ImageNet~\cite{deng2009imagenet} for image classification. Then it is used as the backbone of two  detectors (RetinaNet~\cite{lin2017focal} and Mask R-CNN~\cite{he2017mask}) for object detection and instance segmentation on COCO dataset~\cite{lin2014microsoft}. As for semantic segmentation, the widely used semantic FPN~\cite{kirillov2019panoptic} on ADE20K~\cite{zhou2019semantic} is adopted. Finally, ablation studies are conducted to verify the effectiveness of each component.

\subsection{Image Classification on ImageNet} 

\noindent\textbf{Settings.} We conduct image classification experiments on the benchmark dataset ImageNet~\cite{deng2009imagenet}, which contains 1.28M training images and 50k validation images from 1000 classes. For a fair comparison, we use the same training strategy as \cite{touvron2021training}. Specially, the model is trained  for 300 epochs with AdamW~\cite{loshchilov2017decoupled} optimizer, whose  learning rate is initialized as  0.001 and declines with a cosine decay strategy.  The batchsize and weight decay are set to 1024 and 0.05, respectively. We use the common data augmentation strategies following \cite{touvron2021training}, containing  Mixup~\cite{zhang2017mixup}, CutMix~\cite{yun2019cutmix} and Rand-Augment~\cite{cubuk2020randaugment}. At the inference phase, the top-1 accuracy on a single crop is reported. To be compatible with the downstream tasks, we use a local window for token-FC and set the window size to 7 empirically. By adjusting the architecture configures, four models (T, S ,M, B) with different parameters and computational costs are developed. Besides, by replacing the FC layer of phase estimation module with a depth-wise convolution, an more efficient architecture is developed and denoted as Wave-MLP-T*. All the experiments are conducted with PyTorch~\cite{paszke2017automatic}  and MindSpore~\cite{mindspore} on NVIDIA V100 GPUs.

\noindent\textbf{Comparison with the existing MLP-like architectures.} Table~\ref{tab-mlp} compares the proposed Wave-MLP with existing vision MLP models proposed recently or currently. Throughput is measured on a V100 GPU following \cite{touvron2021training, liu2021Swin}\footnote{Note that AS-MLP~\cite{lian2021mlp} reports throughput under the mixed precision mode~(mixed FP16 and FP32). For a fair comparison with the existing models, we remeasure it with pure FP32 following \cite{touvron2021training, liu2021Swin}}. The family of Wave-MLP achieves a better trade-off between the computational cost and accuracies than the existing methods. For example, our Wave-MLP-M model achieves 83.4\% top-1 accuracy with only 7.9G FLOPs, which shows a large superiority to ResMLP-B24~\cite{touvron2021resmlp}~(81.0\% accuracy with 23.0G FLOPs). Compared with the SOTA MLP architecture CycleMLP~\cite{chen2021cyclemlp}, Wave-MLP also achieves higher accuracies with similar parameters and FLOPs, \eg, Wave-MLP-T achieve a accuracy of  80.6\%, much higher than that of CycleMLP-B1 with 78.9\% accuracy. It shows that equipping each token with the phase information can well capture the relationship between varying tokens and fixed weights to improve the performance of MLP architecture.

\noindent\textbf{Comparison with SOTA models on ImageNet.} We further compare the proposed Wave-MLP  with typical CNN and transformer architectures on ImageNet in Table~\ref{tab-sota}. Compared with  Swin Transformer~\cite{liu2021Swin}, our Wave-MLP achieves higher performance with fewer parameters and computational costs. For example, with 4.5G FLOPs, Wave-MLP-S achieves 82.6\% top-1 accuracy, wihch significantly superior to Swin-T with 81.3\% accuracy. Its trade-off between computational cost and accuracy also suppresses the typical CNN architectures such as RegNetY and ResNet18. The superiority of Wave-MLP implies that the simple MLP architecture has a large potential and modulating the token aggregating process with phase term can exploit it adequately.

\subsection{Object Detection on COCO}
\noindent\textbf{Settings.} We further investigate the  proposed Wave-MLP architecture on the object detection and instance segmentation tasks. The experiments are  conducted on the COCO 2017 dataset~\cite{lin2014microsoft}, which contains 118k training images and 5k validation images. Wave-MLP is used as the backbone and embedded into two prevalent detectors, RetinaNet~\cite{lin2017focal} and Mask R-CNN~\cite{he2017mask}. For a fair comparison, we follow the training recipe in \cite{wang2021pyramid} and train the model with AdamW~\cite{loshchilov2017decoupled} optimizer for 12 epochs~(1$\times$ training scheduler). The batchsize is set to 16 and initial learning to 0.0001. The backbones are initialized with the pre-trained weights on ImageNet while other layers are initialized with Xavier~\cite{glorot2010understanding}.

\noindent\textbf{Results.}  Table~\ref{tab-coco} compares the object detection results with different architectures as the backbone. For both RetinaNet and Mask R-CNN, the proposed Wave-MLP achieves obviously higher performance compared with the existing models. For example, With RetinaNet 1$\times$, Wave-MLP-T achieves 40.4\% AP with only 25.3M parameters and 196.3G FlOPs, which is  higher than  CycleMLP-B1~(38.6 AP) with similar model size by 1.8 AP. When using Mask R-CNN as the detector, the performance improvements are also significant. Compared with Swin-T of 42.2 box AP and 39.1 mask AP with 47.8M parameters and 264.0G FLOPs, our Wave-MLP-S achieves significantly higher performance (44.0 box AP and 40.0 mask AP) with  fewer parameters (47.0M) and lower computational cost (250.3G).

\subsection{Semantic Segmentation on ADE20K}

\noindent\textbf{Settings.} The experiments for the semantic segmentation task are conducted on the challenging ADE20K dataset~\cite{zhou2019semantic}, which contains 25k images from 150 semantic categories, 20k for training, 2k for validation and 3k for testing. Following \cite{wang2021pyramid}, we combine the proposed Wave-MLP architecture with the widely used Semantic FPN~\cite{kirillov2019panoptic} approach. With the pre-trained weights on ImageNet, the model is fine-tuned for 40k iterations with AdamW~\cite{loshchilov2017decoupled} optimizer and the batchsize is set to 32. The initial learning rate is 0.0001 and decays with the polynomial schedule (a power of 0.9). The images are randomly resized and cropped to $512\times 512$ for training and rescaled to have a shorter side of 512 for testing. The FLOPs are tested with 2048$\times$512 input.

\begin{table}
	
	\centering
	\small 
	\caption{The effectiveness of phase information.} 
	\label{tab-phase}
	\vspace{-2mm}
	\begin{tabular}{l|c|c|c}
		\toprule[1.5pt]
	
		Mode& Params. & FLOPs   &  Top-1 accuracy (\%) \\ \midrule 
	
		No phase    &  15M   &2.1G&  78.8   \\
		Static phase &15M &2.1G& 79.3  \\ 
		Dynamic phase & 15M&2.1G& 80.1 \\ 
		\bottomrule[1.5pt]
	\end{tabular}
	\vspace{-2mm}
\end{table}

\begin{table} 
	\centering
	\small 
	\caption{The formulation of phase estimation module.} 
	\label{tab-form}
	\vspace{-2mm}

	\begin{tabular}{l|c|c|c}

		\toprule[1.5pt]
		size& Params. & FLOPs   &  Top-1 accuracy (\%) \\ \midrule 
	
		Baseline &15M& 2.1G  & 78.8 \\
		Identity & 15M &2.1G&  79.3 \\  
		Depth-wise &15M& 2.1G  & 80.1 \\
		Channel-FC &17M& 2.4G  & 80.6 \\ 
		\bottomrule[1.5pt]
	\end{tabular}
\vspace{-4mm}
\end{table}

\begin{table}
	\centering
	\small 
	\caption{The size of window for aggregating tokens.} 
	\vspace{-2mm}
	\label{tab-size}
	
	\begin{tabular}{l|c|c|c}
		\toprule[1.5pt]
		Size& Params. & FLOPs   &  Top-1 accuracy (\%) \\ \midrule 
	
		3         &15M& 2.1G  & 79.7 \\
		5         &15M& 2.1G  & 79.8 \\
		7 &15M &2.1G& 80.1  \\ 
		All &16M &2.3G&80.0  \\ 
		\bottomrule[1.5pt]
	\end{tabular}
\vspace{-4mm}
\end{table}

\noindent\textbf{Results.} The results of different models for semantic segmentation are shown in Table~\ref{tab-ade}. Under different configures of parameters and computational costs, Wave-MLP outperforms the existing models consistently. Compared with the transformer-based model such as PVT, the model show a large superiority, \eg, 4.6\% mIoU gap between Wave-MLP-S (44.4\% mIoU) and PVT-Tiny (39.8\% mIoU). It also suppress the CycleMLP-B2 model with 42.4\% mIoU and Swin-S with 41.5\%. We infer that modulating the aggregating process of different tokens with the phase term can capture more detailed information and thus enhance the semantic segmentation results.

\subsection{Ablation Studies} 
\label{sec-abl}
For better understanding the proposed method, we investigate the effectiveness of each component via ablation studies. The experiments are conducted on ImageNet with the Wave-MLP-T* model.
 
\noindent\textbf{The effectiveness of phase information.} The phase plays a vital role in aggregating the information of different tokens, whose effectiveness is investigated in Table~\ref{tab-phase}.  Without the phase information~(`No phase'), the model's performance is obviously inferior  compared with others, with only 78.8\% top-1 accuracy.  The  proposed 'dynamic phase' flexibly generates phases and modulate the aggregating process for each input instance, which achieves much better performance~(\eg, 80.1\% top-1 accuracy).

\noindent\textbf{The formulation of phase estimation module.} The phase estimation module generates phases for different inputs, which can be implemented with different formulations. We investigate three simple formulations, depth-wise convolution, channel-FC and identity projection. The identity projection directly  copies the input feature instead of estimating the phase, incurring poor performance~(\ie, 79.3\%). 
The depth-wise  convolution and channel-FC can achieve high accuracy improvement compared with the baseline~(\eg, 1.3\% and 1.8\%), implying they can capture the phase information well for aggregating tokens. Using channel-FC achieves higher performance than the depth-wise convolution, but also increase the computational cost slightly.   

\noindent\textbf{The size of window for aggregating tokens.} To be compatible with dense prediction tasks (\eg, object detection and semantic segmentation) with varying sizes of input images, we restrict that the token-FC only aggregates features within a local window, and Table~\ref{tab-size} investigates the impact of window size. Changing window size from 3 to 7, the top-1 accuracies increase accordingly. `All' denotes that the token-FC connects all the tokens in a layer, which achieves similar performance with window size 7. However, its parameter configure is corrected to the size of input image, and thus is infeasible in dense prediction tasks such as object detection and semantic segmentation.  
\begin{figure}[t] 
	\centering
	\small
	\includegraphics[width=0.8\linewidth]{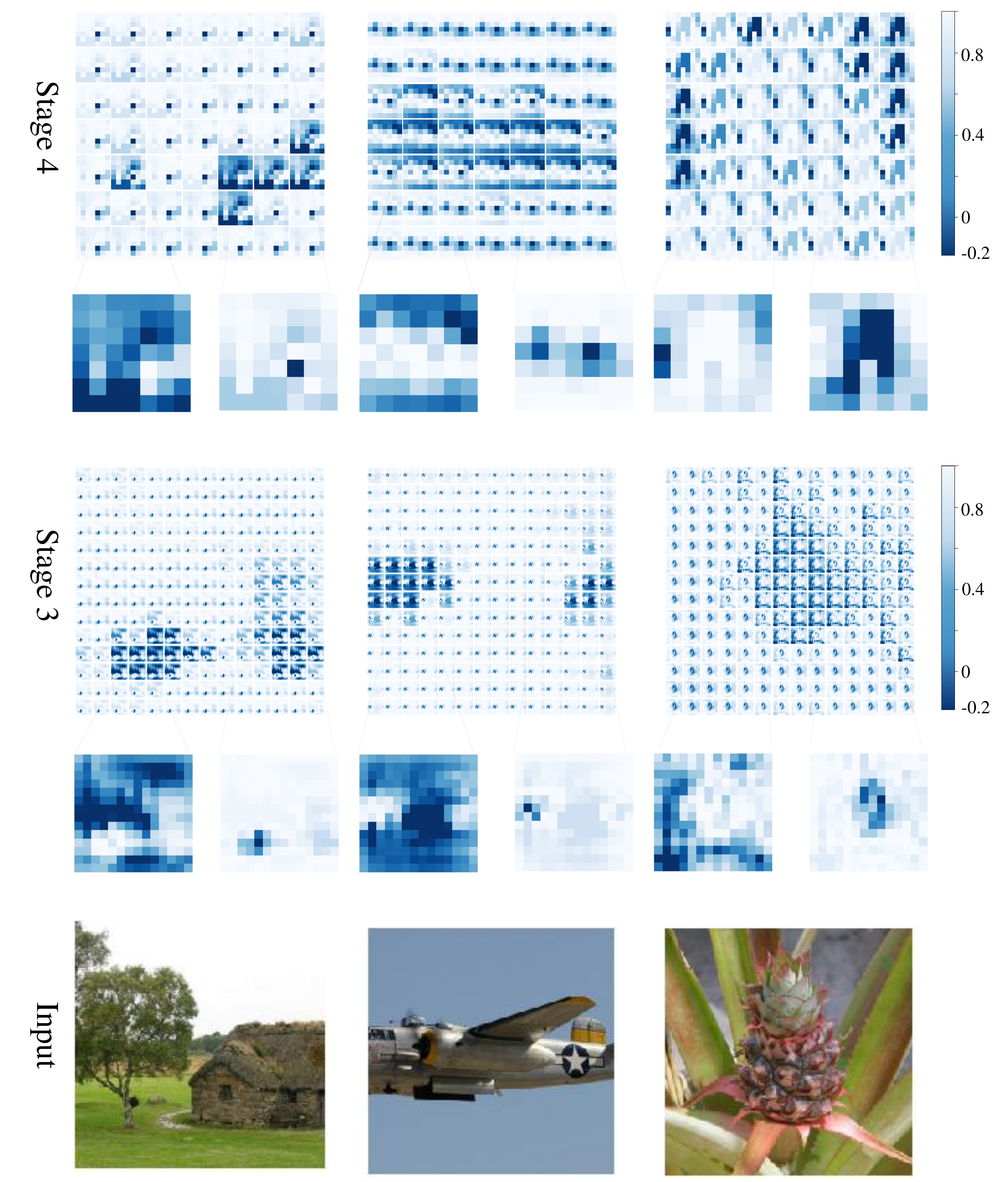}		
	\caption{Visualization of the phase difference between tokens.}
	\label{fig-theta}
	\vspace{-6mm}	
\end{figure}

\noindent\textbf{Visualization.} The phase difference between two tokens ($|\btheta_j-\btheta_i|$) directly affects the aggregating process as analyzed in Section~\ref{sec-patm}~(Eq.~\ref{eq-amp2}, \ref{eq-phs2}). In order to have an intuitive understanding, we show the cosine value of phase difference of the 3rd and 4th stages in Figure~\ref{fig-theta}. Take the visualized figure of the 1st image and the 4th stage for example, the $7\times7$ values in the $(i,j)$-th patch denote the phase differences between the $(i,j)$-th token and all the $7\times7$ tokens. From the figure, we can see that tokens with similar contents tend to have close phases and then enhanced by each other. For example, in the first image, a token describing the 'house' has a closer phase with another token of the `house' than that of the sky (magnifying parts in the figure). The phase difference of different tokens also varies \wrt different input images depending on the image contents.

\vspace{-2mm}
\section{Conclusion}

This paper proposes a Wave-MLP architecture for vision tasks, which takes each token as a wave with both amplitude and phase information. Amplitude is the original real-value feature and the phase modulates  relationship between the varying tokens and fixed weights in MLP. With the dynamically produced phase, the tokens are aggregated according to their varying contents from different input images. Extensive experiments show that the proposed Wave-MLP suppresses the existing MLP-like architectures and can also be used as a strong backbone for the dense prediction tasks such as object detection and semantic segmentation. In the future, we will further explore the potential of MLP-like architectures on more diverse tasks.
\label{sec-con}

\noindent\textbf{Acknowledgment.} This work is supported by National Natural Science Foundation of China under Grant No.61876007, Australian Research Council under Project DP210101859 and the University of Sydney SOAR Prize.

{\small
\bibliographystyle{ieee_fullname}
\bibliography{egbib}
}

\twocolumn[{%
	\renewcommand\twocolumn[1][]{#1}%
	\maketitle
	\begin{center}
		\centering
		\captionsetup{type=table}
		\small
		\caption{Detailed architecture specifications of Wave-MLP. `Dimension' and `expansion' denote the dimension of feature and expand ratio, respectively. H and W are the height and width of input image. FLOPs is calculated with input size of 224$\times$224.}
		\label{table-archconf}

		\begin{tabular}{c|c|c|c|c|c} 
			\toprule[1.5pt]
			& Output size & Wave-MLP-T  & Wave-MLP-S & Wave-MLP-M & Wave-MLP-B \\
			\hline
			
			\multirow{1}{*}{stage 1} & \multirow{1}{*}{$\frac{H}{4}\times \frac{W}{4}$} 
			
			& $\begin{bmatrix}\text{dimension = 64}\\\text{expansion = 4}\end{bmatrix}$ $\times$ 2   & $\begin{bmatrix}\text{dimension = 64}\\\text{expansion = 4}\end{bmatrix}$ $\times$ 2    & $\begin{bmatrix}\text{dimension = 64}\\\text{expansion = 8}\end{bmatrix}$ $\times$ 3   & $\begin{bmatrix}\text{dimension = 96}\\\text{expansion = 4}\end{bmatrix}$ $\times$ 2   \\
			\hline
			\multirow{1}{*}{stage 2}  & \multirow{1}{*}{$\frac{H}{8}\times \frac{W}{8}$} &
			$\begin{bmatrix}\text{dimension = 128}\\\text{expansion = 4}\end{bmatrix}$ $\times$ 2  & $\begin{bmatrix}\text{dimension = 128}\\\text{expansion = 4}\end{bmatrix}$ $\times$ 3 & $\begin{bmatrix}\text{dimension = 128}\\\text{expansion = 8}\end{bmatrix}$ $\times$ 4 & $\begin{bmatrix}\text{dimension = 192}\\\text{expansion = 4}\end{bmatrix}$ $\times$ 2 \\
			\hline
			\multirow{1}{*}{stage 3}  & $\frac{H}{16}\times \frac{W}{16}$& $\begin{bmatrix}\text{dimension = 320}\\\text{expansion = 4}\end{bmatrix}$ $\times$ 4 & $\begin{bmatrix}\text{dimension = 320}\\\text{expansion = 4}\end{bmatrix}$ $\times$ 10 & $\begin{bmatrix}\text{dimension = 320}\\\text{expansion = 4}\end{bmatrix}$ $\times$ 18  & $\begin{bmatrix}\text{dimension = 384}\\\text{expansion = 4}\end{bmatrix}$ $\times$ 18 \\
			\hline 
			\multirow{1}{*}{stage 4}  & $\frac{H}{32}\times \frac{W}{32}$& $\begin{bmatrix}\text{dimension = 512}\\\text{expansion = 4}\end{bmatrix}$ $\times$ 2 & $\begin{bmatrix}\text{dimension = 512}\\\text{expansion = 4}\end{bmatrix}$ $\times$ 3 & $\begin{bmatrix}\text{dimension = 512}\\\text{expansion = 4}\end{bmatrix}$ $\times$ 3  & $\begin{bmatrix}\text{dimension = 768}\\\text{expansion = 4}\end{bmatrix}$ $\times$ 2 \\ \hline
			\multicolumn{2}{c|}{\# Parameters}&17M&30M&44M&63M\\ \hline
			\multicolumn{2}{c|}{FLOPs}&2.4G&4.5G&7.9G&10.2G\\ \bottomrule[1.5pt]
		\end{tabular}
	\end{center}

	\begin{center}
		\centering
		\captionsetup{type=table}
		
		\small
		\caption{Results of object detection and instance segmentation on COCO val2017. The Mask R-CNN model trained with 3$\times$ schedule and multi-scale training strategy~\cite{he2017mask} is used as the detector.}
		\label{tab-coco3}

		\begin{tabular}{l| c|lccccc|clcccc}
			
			\toprule[1.5pt]
			
			Backbone& Params. / FLOPs& AP$^{\rm b}$ &AP$_{50}^{\rm b}$ &AP$_{75}^{\rm b}$&AP$_{S}^{\rm b}$&AP$_{M}^{\rm b}$& AP$_{L}^{\rm b}$ &AP$^{\rm m}$ &AP$_{50}^{\rm m}$ &AP$_{75}^{\rm m}$&AP$_{S}^{\rm m}$&AP$_{M}^{\rm m}$&AP$_{L}^{\rm m}$\\ \hline
			
			ResNet18~\cite{he2016deep}       & 31.2M / 207.3G & 36.9 & 57.1 & 40.0 &-&-&-& 33.6 & 53.9 & 35.7&-&-&- \\
			PVT-Tiny~\cite{wang2021pyramid}  & 32.9M / 208.1G & 39.8 & 62.2 & 43.0 &-&-&-& 37.4 & 59.3 & 39.9&-&-& -\\
			Wave-MLP-T                      & 25.3M / 196.3G&\textbf{44.1}&66.0&48.2&28.4&47.6&55.9&\textbf{40.1}&63.1&43.2&24.3&43.5&53.2\\
			\hline
			ResNet50~\cite{he2016deep}       & 44.2M / 260.1G & 41.0 &  61.7 & 44.9 &-&-&-& 37.1 & 58.4 & 40.1&-&-&- \\
			PVT-Small~\cite{wang2021pyramid}  & 44.1M / 245.1G & 43.0 & 65.3 & 46.9 &-&-&-& 39.9 &62.5 & 42.8&-&-& -\\
			Wave-MLP-S                          & 37.1M / 231.3G&\textbf{45.5}&66.9&49.3&29.4&48.7&58.7&\textbf{41.0}&64.2&44.0&25.0&44.2&54.7\\
			\hline
			ResNet101~\cite{he2016deep} &63.2M / 336.4G& 42.8 & 63.2 &-&-&-& 47.1 & 38.5& 60.1& 41.3&-&-&-\\
			PVT-Medium &63.9M / 301.7G & 44.2 & 66.0 & 48.2 &-&-&-& 40.5 & {63.1} & 43.5&-&-&-\\
			PVT-Large &71.1M / 345.7G & 44.5 & 66.0 & 48.3 &-&-&-& 40.7 & {63.4} & 43.7&-&-&-\\
			Wave-MLP-M  &49.4M / 291.3G &\textbf{46.3}&67.8&50.3&29.5&49.3&60.3&\textbf{41.5}&65.2&44.1&24.9&44.7&55.6\\
			\bottomrule[1.5pt]
		\end{tabular}
		
	\end{center}
}]

\section{Detailed Architectures}
\label{sec-con}

Table~\ref{table-archconf} shows the detailed  specifications of the proposed Wave-MLP architecture. To get hierarchical features, we split the whole model into four stages, and reduce the size of feature map stage-wisely.  The Wave-MLP family contains four models with different parameters and computational costs by adjusting the depths and widths of architecture specifications, which are denoted as Wave-MLP-T, Wave-MLP-S, Wave-MLP-M, and Wave-MLP-B, sequentially.
From Wave-MLP-T to Wave-MLP-B, the number of parameters varies from 17M to 63M, and FLOPs varies from 2.4G to 10.2G.

\section{More Experiments}

For the object detection and instance segmentation tasks on COCO~\cite{lin2014microsoft}, we further  train  Mask R-CNN models with 3$\times$  schedule and  multi-scale training strategy~\cite{he2017mask}. The results of different backbone are shown in Table~\ref{tab-coco3}. Compared with other backbones, the proposed Wave-MLP achieves much higher performance. For example, our Wave-MLP-T achieves 44.1 box AP and 40.1 mask AP with 25.3M parameters and 196.3G FLOPs, which is significantly superior to the PVT-Tiny model with 39.8 box AP, 37.4 mask AP, 32.9M parameters and 208.1G FLOPs.

\end{document}